\pdfoutput=1
\documentclass[11pt, table, xcdraw]{article}

\usepackage[final]{acl}

\usepackage{times}
\usepackage{latexsym}

\usepackage[T1]{fontenc}

\usepackage[utf8]{inputenc}

\usepackage{microtype}

\usepackage{inconsolata}

\usepackage{graphicx}

\usepackage{booktabs}

\usepackage{subfigure} 

\usepackage{multirow}
\usepackage{multicol}
\usepackage{amssymb} 
\usepackage{colortbl} 
\usepackage{amsmath, amssymb, amsthm}
\usepackage{float}
\usepackage{arydshln}
\usepackage{pifont}
\usepackage{enumitem}

\usepackage{makecell}
\usepackage[normalem]{ulem}
\author{Shuyu Guo\thanks{Work performed during a visit to Leiden University.} \\
  Shandong University \\
  Qingdao, China \\
  \texttt{guoshuyu225@gmail.com} \\\And
  Shuo Zhang \\
  Bloomberg \\
  London, United Kingdom \\
  \texttt{szhang611@bloomberg.net} \\\And
  Zhaochun Ren\thanks{Corresponding Author.} \\
  Leiden University \\
  Leiden, The Netherlands \\
  \texttt{z.ren@liacs.leidenuniv.nl} \\}

\title{Enhancing RAG Efficiency with Adaptive Context Compression}

\begin{document}
    \maketitle

    \begin{abstract}
Retrieval-augmented generation (RAG) enhances large language models (LLMs) with external knowledge but incurs significant inference costs due to lengthy retrieved contexts. While context compression mitigates this issue, existing methods apply fixed compression rates—over-compressing simple queries or under-compressing complex ones. We propose Adaptive Context Compression for RAG (ACC-RAG), a framework that dynamically adjusts compression rates based on input complexity, optimizing inference efficiency without loss of accuracy. ACC-RAG combines a hierarchical compressor (for multi-granular embeddings) with a context selector to retain minimal sufficient information, akin to human skimming. Evaluated on Wikipedia and five QA datasets, ACC-RAG outperforms fixed-rate methods and unlocks >4× faster inference versus standard RAG while maintaining or improving accuracy.
\end{abstract}
    \section{Introduction}
\label{sec:intro}

Large Language Models (LLMs) are pre-trained on massive datasets, encoding vast knowledge within billions of parameters. While they excel at many tasks, their parametric knowledge often falls short for knowledge-intensive applications. Retrieval-Augmented Generation (RAG) addresses this limitation by extending the model's knowledge boundaries through external context retrieval~\cite{DBLP:journals/corr/abs-2312-10997}. However, integrating lengthy retrieved contexts into prompts increases inference costs and may exceed LLMs' context window limits~\cite{autocompr}.

\begin{figure}[t]
    \centering
    \includegraphics[width=\columnwidth]{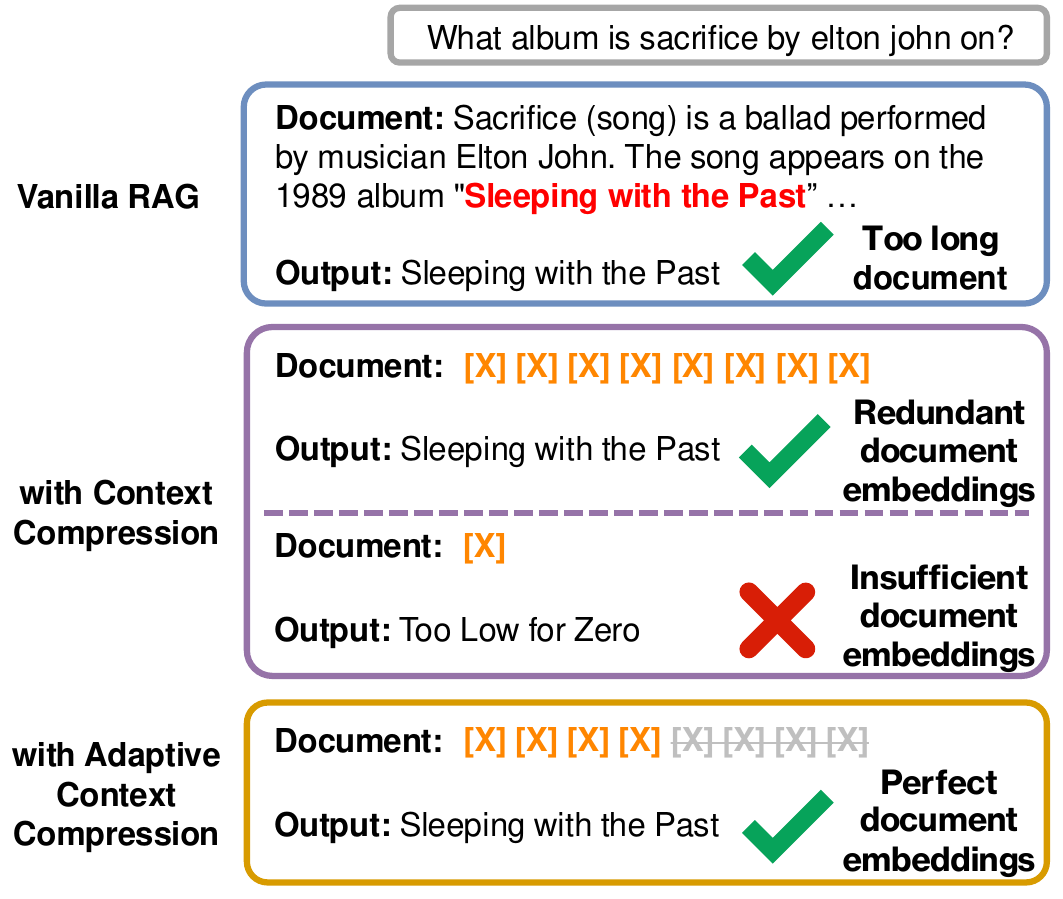}
    \caption{An example of a retrieval-augmented model with different context compression methods.}
    \label{fig:main-b}
\end{figure}

Context compression mitigates this issue by transforming long contexts into shorter input sequences. Existing methods fall into two categories: (1) Lexical-based compression, which reduces input length by preserving key tokens~\cite{filc} or generating summaries~\cite{recomp}; and (2) Embedding-based compression, which encodes text into dense embeddings~\cite{icae} for inference.
Embedding-based methods have been proven more efficient and effective~\cite{xrag}, typically employing a compressor trained in two phases: first through pre-training (e.g., via autoencoding or language modeling) to preserve contextual information~\cite{autocompr}, followed by fine-tuning (e.g., with instruction follow-up or self-distillation) to adapt to downstream tasks~\cite{cocom}.
Existing embedding-based approaches fix the compression rate (i.e., token-to-embedding ratio), leading to trade-offs: high rates risk losing essential information, while low rates retain redundancies (Figure~\ref{fig:main-b}). Additionally, inconsistent evaluation benchmarks—varying training data scales and tasks across baselines—hinder fair comparisons.

We propose an adaptive context compression framework for RAG that dynamically adjusts the number of compressed embeddings during inference. The framework decouples offline compression (fixed-rate hierarchical embeddings) from online selection (dynamic embedding feeding, halted once sufficient context is accumulated), mimicking human selective reading.
We establish a unified benchmark (Wikipedia corpus + five QA datasets) to ensure fair evaluation. Our method outperforms other compression techniques in effectiveness and efficiency, matches or exceeds standard RAG accuracy on four datasets with $4\times$ faster inference, and demonstrates significant potential for adaptive compression via compressor analysis.

In summary, our contributions include: (1) A novel adaptive context compression
framework for RAG, improving inference efficiency while maintaining retrieval augmentation
benefits. (2) Training strategies for a hierarchical compressor and adaptive
selector, enabling multi-granularity compressed embedding generation and automatic
context selection.
(3) A unified benchmark for context compression methods, resolving evaluation biases from inconsistent training data and task setups in prior works. (4) Experimental results showing our method as the best under context compression, with comparable results to regular RAG but with improved efficiency. (5) Further analysis highlighting the potential of adaptive context compression.

\section{Related Work}
\label{sec:related_work}

\subsection{Retrieval-Augmented Generation}
Retrieval-augmented generation (RAG) improves model performance across tasks~\cite{DBLP:journals/corr/abs-2312-10997, DBLP:conf/acm/AsaiMZC23}, including language modeling~\cite{DBLP:conf/acl/MinSL0YHZ23,DBLP:conf/nips/Wang0CLYGW23}, question answering~\cite{DBLP:conf/nips/LewisPPPKGKLYR020, DBLP:conf/naacl/ShiMYS0LZY24,DBLP:conf/sigir/XuCGWDWL24,DBLP:conf/acl/Xiong0LZ24}, machine translation~\cite{DBLP:conf/iclr/KhandelwalFJZL21,DBLP:conf/emnlp/ChengGL0022}, code generation~\cite{DBLP:conf/emnlp/Zhang0YC23,DBLP:conf/emnlp/ZhangCZKLZMLC23}, and more. Various approaches have been proposed, including end-to-end optimization of model and retriever~\cite{DBLP:conf/icml/GuuLTPC20}, enhanced integration with non-parametric knowledge~\cite{DBLP:conf/icml/BorgeaudMHCRM0L22,cheng-etal-2023-decouple}, improved alignment between model and retriever~\cite{DBLP:conf/naacl/ShiMYS0LZY24,DBLP:conf/iclr/Lin0CSL00KSLZY24}, and the integration of a self-reflection mechanism~\cite{selfmem,selfrag}. However, most focus on improving effectiveness, neglecting efficiency degradation from incorporating external knowledge.

\subsection{Context Compression}
Context compression aims to shorten model input text, improving inference speed. It's a key solution to mitigate efficiency degradation in RAG systems~\cite{ccs}. Methods fall into two categories: lexical-based, which reduce input context length by extracting tokens~\cite{filc} or summarizing context~\cite{recomp}, and embedding-based, which encode the context into embeddings to replace textual input~\cite{gist}. Embedding-based methods currently outperform lexical-based due to their flexible information storage.
Embedding-based methods have shown strong performance. 
\citet{rmt} segmented long context into chunks and encoding their information into memory tokens. 
\citet{ftr} introduced context compression and applied it to toxicity reduction.
\citet{gist} proposed using gist tokens to generalize context compression. 
\citet{autocompr} compressed long contexts into summary vectors. 
\citet{icae} employed autoencoding and language modeling objectives to pretrain compressors. 
\citet{xrag} projected retrieval embeddings into context embeddings to achieve extreme compression. 
\citet{cocom} jointly optimized the compressor and decoder, extending the compression capability to multiple documents. 
Notably, existing methods rely on static compression strategies that cannot adapt to varying question difficulty. Additionally, differences in training data scales and task mixtures across studies impede fair comparison. Our work resolves both limitations.
    \section{Adaptive Context Compression for RAG}
\label{sec:methods}

\begin{figure*}[t]
    \centering
    \includegraphics[width=1.8\columnwidth]{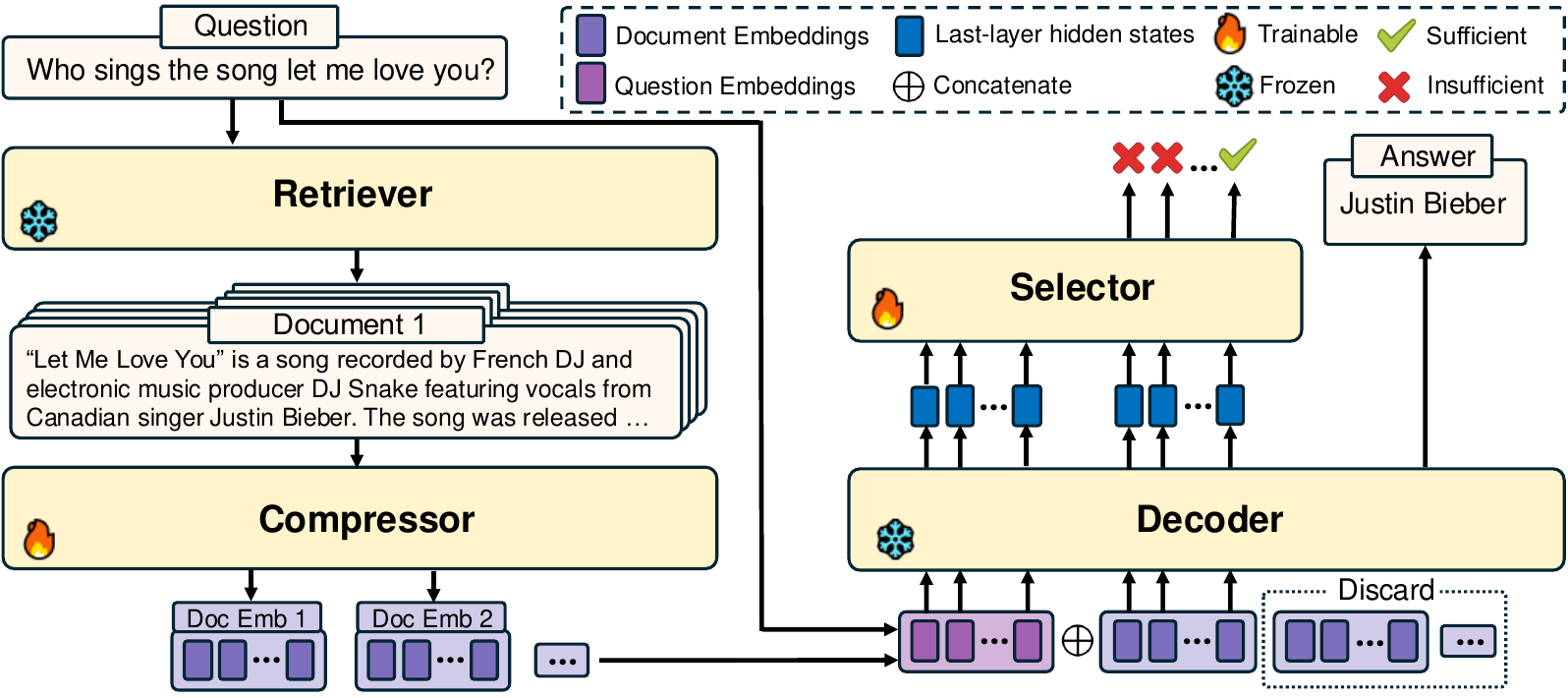}
    \caption{Overview of Adaptive Context Compression for RAG pipeline. The workflow
    is as follow: Context Retrieval $\xrightarrow{}$ Hierarchical Compression
    $\xrightarrow{}$ Adaptive Context Selection $\xrightarrow{}$ Response
    Generation.}
    \label{figure:framework}
\end{figure*}

A RAG framework typically consists of a retriever $\mathcal{R}$ and a decoder
$\mathcal{D}$. The retriever builds a search index $I$ containing dense representations
of all documents from a collection. For a given query $Q$, $\mathcal{R}$ retrieve
the top-$k$ relevant documents $D = \{d_{1}, d_{2}, \ldots, d_{k}\}$ by matching
$Q$'s embedding against $I$ via similarity search. The query $Q$ and documents
$D$ are concatenated and fed into the decoder $\mathcal{D}$ to generate a
response $R$. This increases the input size to the decoder, reducing inference
efficiency as $|D| \gg |Q|$.

Existing embedding-based compression methods employ a fixed-rate compressor to convert documents into embedding sequences, which are subsequently fed into the decoder for answer generation instead of the original documents. However, this fixed-ratio approach lacks adaptability to query complexity. The ACC-RAG framework introduces a \textit{hierarchical compressor} $\mathcal{C}$
C and an \textit{adaptive selector} $\mathcal{S}$ working together to support dynamic compression. The former encodes documents into multi-granular document embeddings, enabling variable information density across segments. The latter progressively feeds embeddings into the decoder and halts once sufficient context is reached, effectively controlling input sequence length. This mechanism allows variable-length decoder inputs, thereby realizing dynamic compression.
The full workflow is illustrated in Figure~\ref{figure:framework}.

\subsection{Hierarchical Compressor}


The compressor $\mathcal{C}$ processes each document $d_{i}$ into a multi-granular embedding sequence. Specifically, the input text is concatenated with $m_{i}= \lfloor L_{i}/\tau \rfloor$ special compression tokens, where $L_{i}$ is the document length and $\tau$ is a fixed compression rate. These compression tokens are encoded, and their final-layer embeddings form the compressed sequence $E_{i}$:

\begin{equation}
E_{i}= \mathcal{C}(d_{i}) = [e_{i}^{(1)}, \ldots, e_{i}^{(m_i)}]
\end{equation}

\subsubsection{Fix-rated Preprocessing}

The compressor employs a fixed compression ratio: higher ratios yield shorter compressed sequences with reduced information retention. The entire compression process operates offline, with all documents pre-processed into their embedding sequences for instant access during retrieval.


\subsubsection{Training Strategy}

The compressor is trained through a widely adopted two-stage process: pretraining and fine-tuning. The former enables the compressor to preserve maximal contextual information, while the latter facilitates adaptation to downstream tasks. During pretraining, the compressor learns to encode text into embeddings through two tasks~\cite{icae}: (1) auto-encoding, where the decoder reconstructs the original text from the compressed embeddings alone, and (2) language modeling, where the decoder generates text continuations conditioned on the embeddings. Formally, given a task instruction $Q$, compressed sequence $E_{c}$, and target response $R = \{r_{1}, \ldots, r_{n}\}$ (either original text or continuation), both tasks optimize the negative log-likelihood (NLL) loss:
\begin{equation}
    \mathcal{L}(\theta_{\mathcal{C}}) = -\sum_{t=1}^{n}\log
    P_{\theta_{\mathcal{D}}}(r_{t}\mid Q, E_{c}, r_{<t})
\end{equation}
Only the compressor parameters $\theta_{\mathcal{C}}$ are updated, while the decoder parameters $\theta_{\mathcal{D}}$ remain frozen.
During fine-tuning, we employ a self-distillation task~\cite{self-dis}, where the original RAG model acts as a teacher to preserve decoder capabilities. Given a query $Q$, context $C$, and target response $R$, the teacher distribution is $P_{\text{teacher}}(r_{t}\mid Q, C, r_{<t})$, and the student distribution is $P_{\text{student}}(r_{t}\mid Q, E_{c}, r_{<t})$. The training minimizes the KL divergence between these distributions:
\begin{equation}
\mathcal{L} = D_{\text{KL}}\left(P_{\text{teacher}} \parallel P_{\text{student}}\right)
\end{equation}
Notably, the fine-tuning approach avoids instruction-following tasks, preventing the compressed embeddings from altering the decoder's original generation pattern.



\subsubsection{Multi-granularity Compression}

In addition to encoding useful information, the compressor must also learn hierarchical information organization. We define granularity $b$ as the starting position of an embedding sequence. The compressed sequence should exhibit varying information densities across subsequences truncated at different granularity levels.
To achieve multi-granular compression, we optimize across multiple granularities during both pretraining and fine-tuning. Formally, given a training granularity sequence $\mathcal{B}= \{b_{1}, \ldots, b_{k}\}$, the training objective combines reconstruction errors at all granularities:
\begin{equation}
    \mathcal{L}(\theta_{\mathcal{C}}) = -\sum_{b \in \mathcal{B}}\sum_{t=1}^{n}\log
    P_{\theta_{\mathcal{D}}}(r_{t}\mid Q, E_{c}^{1:b}, r_{<t})
\end{equation}
where $E_{c}^{1:b}= [e_{1}, \ldots, e_{b}]$. 
This method ensures $\mathcal{C}$ allocates crucial information early and maintains complementary details in later positions.

\subsection{Adaptive Selector}
The adaptive selector $\mathcal{S}$ dynamically controls decoder input by progressively accumulating and verifying context embeddings during inference. 
The query is encoded into a query embedding sequence $E_{q} = \mathcal{D}_{\text{embed}}(q)$, while the context embedding sequence is obtained by concatenating all compressed document sequences in retrieval order: $E_{c} = \mathrm{Concat}(E_{1}, \ldots, E_{k})$. 
The selector iterates through granularity levels from the inference granularity sequence $\mathcal{B} = \{b_{1}, \ldots, b_{k}\}$. At each step $t$:

\begin{enumerate}[leftmargin=*, noitemsep, topsep=2pt]

    \item Extract the subsequence $E_{1:b_t}= [e_{1}, \ldots, e_{b_t}]$ from $E_{c}$,
        then concatenate with $E_{q}$ to construct the current input:
        \begin{equation}
            X_{t}= \mathrm{Concat}(E_{q}, E_{1:b_t})
        \end{equation}

    \item Feed $X_{t}$ into the decoder $\mathcal{D}$ to obtain the last layer hidden
        states:
        \begin{equation}
            H_{t}= \mathcal{D}(X_{t})
        \end{equation}

    \item Feed $H_{t}$ into the selector $\mathcal{S}$ to verify if the context information
        is sufficient:
        \begin{equation}
            \mathcal{S}(H_{t}) =
            \begin{cases}
                1 & \text{sufficient}   \\
                0 & \text{insufficient}
            \end{cases}
        \end{equation}
\end{enumerate}

This process continues until $\mathcal{S}(H_{t}) = 1$ or $t = k$. The final input
$X$ is determined based on the termination state:
\begin{equation}
    X =
    \begin{cases}
        X_{t}                         & \mathcal{S}(H_{t}) = 1 \\
        \mathrm{Concat}(E_{q}, E_{c}) & t=k
    \end{cases}
\end{equation}

\subsubsection{Training Data Synthesis}
The selector $\mathcal{S}$ is trained on synthesized decision tuples. Given an
instruction $I$, context $C$, and gold response $R$, for each granularity $b \in
\mathcal{B}$, the hidden states $H_{b}$ and generated responses $R_{b}$ are
obtained through inference with the decoder, while the label $y_{b}$ is derived by
evaluating $R_{b}$ against $R$:
\begin{equation}
    \begin{aligned}
        H_{b} & = \mathcal{D}(I, E_{c}^{1:b})              \\
        R_{b} & = \mathcal{D}_{\text{gen}}(I, E_{c}^{1:b}) \\
        y_{b} & = \text{eval}(R, R_{b})
    \end{aligned}
\end{equation}
The training data $x_{i}$ consists of pairs:
\begin{equation}
    x_{i}= \left\{(H_{b}, y_{b}) \mid \forall b \in \mathcal{B}\right\}
\end{equation}
A fixed decoder requires only a unified selector, as the generation mode remains
the same.

\subsubsection{Reinforcement Learning}
The selector is trained using reinforcement learning, optimized via policy gradient
with trajectory-based rewards. During inference, we stop adding context embeddings
and generate the response once the selector predicts "sufficient". For each
episode with granularities $b_{1}< \cdots < b_{k}$:
\begin{equation}
    \begin{aligned}
        \text{Trajectory:} & \quad \tau = (s_{1}, a_{1}), \ldots, (s_{T}, a_{T})                               \\
        \text{State:}      & \quad s_{t}= \mathrm{Concat}(H_{I}, H_{b_t})                                      \\
        \text{Action:}     & \quad a_{t}= \begin{cases}1 & \text{sufficient} \\ 0 & \text{continue}\end{cases}
    \end{aligned}
\end{equation}
where $H_{I}$ and $H_{b_t}$ refer to the hidden states of the instruction $I$ and
the truncated embedding sequence $E_{c}^{1:b_t}$, respectively.

The trajectory terminates when the action is 1 or all granularities are
traversed. The reward is:
\begin{equation}
    R(\tau) =
    \begin{cases}
        +1 & \text{if }\exists t: a_{t}= 1 \land y_{b_t}= 1 \\
        -1 & \text{otherwise}
    \end{cases}
\end{equation}
The policy $\pi_{\theta}$ parameters are updated via the REINFORCE algorithm~\cite{ref}:
\begin{equation}
    \nabla_{\theta}J(\theta) = \mathbb{E}_{\tau \sim \pi_\theta}\left[ R(\tau) \sum
    _{t=1}^{T}\nabla_{\theta}\log \pi_{\theta}(a_{t}\mid s_{t}) \right]
\end{equation}
where $\pi_{\theta}$ shares parameters with $\mathcal{S}$.

    \section{Experimental Setup}
\label{sec:experimental_setup}

\subsection{Datasets}
\label{subsec:datasets}
To ensure a fair comparison of different compression methods,
we train and evaluate all methods on unified datasets.
For the retrieval corpus, we use the Wikipedia corpus from Dec. 20, 2018,
standardized by \citet{dpr} using the preprocessing from \citet{chen-etal-2017-reading}.
The corpus is split into multiple, disjoint text blocks of 128 tokens as
documents, resulting in 21,015,324 documents in the end.
For fine-tuning, we conduct experiments on five commonly used open-domain QA benchmarks,
namely Natural Questions (NQ)~\cite{nq}, TriviaQA~\cite{trivalqa}, WebQuestions
(WQ)~\cite{web}, CuratedTREC (TREC)~\cite{trec}, and SQuAD v1.1~\cite{squad}.
All question-answer pairs are ensured to have supporting documents available in
the retrieval corpus.
For pretraining, we select all supporting documents from the training set of QA
and randomly sample additional documents from the retrieval corpus, resulting in
a total of 1 million documents for training.

\subsection{Baselines}
We compare our method with two categories of compression approaches: \emph{plug-in
methods} and \emph{full-tuning methods}. \emph{Plug-in methods} train additional
compressors without modifying the decoder parameters, resulting in low training costs
and preserving the original performance of the decoder. Baseline methods in this
category include:

\textbf{ICAE}~\cite{icae}: Pretrains the comprossor with same task as ours but
finetunes through instruction-following. We implemented ICAE models with multiple
compression rates for evaluation.

\textbf{xRAG}~\cite{xrag}: Maps retrieval embeddings into compression embeddings
for extreme compression. Performs only auto-encoding during pretraining and
combines instruction-following with self-distillation during fine-tuning.
%

\emph{Full-tuning methods} adopt an end-to-end approach, jointly training the
compressor and decoder for better alignment and performance, but at increased training
costs and changes to the original decoder performance. Baseline methods in this
category include:

\textbf{Autocompressor}~\cite{autocompr}: Trains the compressor only with
language modeling. We use the pretraining-only COCOM model as a simplified version.

\textbf{COCOM}~\cite{cocom}: Similar to ICAE but jointly optimizes the compressor
and decoder.


Our method is implemented solely as a plug-in method, offering lower cost,
better scalability, and strong performance. We use the same backbone LLM for all
methods and apply identical LoRA parameters for LoRA-based fine-tuning.

\subsection{Evaluation Metrics}
We evaluate model performance from two dimensions: effectiveness and efficiency.
For effectiveness, we use the Match (M) metric, which measures whether the
reference answer appears in the model’s generated output. We do not use the Exact
Match (EM) metric due to verbose responses from LLMs affecting its calculation.
As for Efficiency, we use First Token Inference Time (FTIT) as the metric.
Unlike widely-adopted Total Inference Time (TIT)~\cite{xrag, cocom}, FTIT focuses on the time to the first token,
isolating the impact of compression methods on efficiency rather than generation
performance.

\begin{table*}[!t]
\centering
\small
\setlength\tabcolsep{4.5pt}
\caption{
The main results between ACC-RAG and other compression methods. 
CR stands for Compression Rate. 
For each dataset, two columns represent Match (M) and First Token Inference Time (FTIT), with the highest M bolded and the second-highest underlined.
\textsuperscript{*} indicates statistical non-significance (p>0.05) with respect to ACC-RAG.
}
\vspace*{-1.5mm}
\label{table:main_results}
\begin{tabular}{c c cc cc cc cc cc cc}
\toprule
\multirow{2}{*}{\textbf{Method}} 
& \multirow{2}{*}{\textbf{CR}}  
& \multicolumn{12}{c}{\textbf{Datasets}}
\\
\cmidrule(lr){3-14}
&
& \multicolumn{2}{c}{\textbf{NQ}}
& \multicolumn{2}{c}{\textbf{TriviaQA}}
& \multicolumn{2}{c}{\textbf{WQ}}
& \multicolumn{2}{c}{\textbf{TREC}}
& \multicolumn{2}{c}{\textbf{SQuAD}}
& \multicolumn{2}{c}{\textbf{Average}}
\\ 
\midrule

LLM & - & 34.79 & 180 & 20.93 & 391 & \textbf{43.36\textsuperscript{*}} & 172 & \textbf{34.29\textsuperscript{*}} & 235 & 20.92 & 210 & 30.86 & 238 \\
RAG & - & \textbf{44.49} & 3264 & \textbf{23.82} & 3529 & 41.78 & 3334 & \underline{33.43} & 3374 & \textbf{47.34} & 3356 & \textbf{38.17} & 3371 \\
\rowcolor[gray]{0.9} 
\multicolumn{14}{c}{\textbf{Full-Tuning Methods}} \\
AutoCompressor & $\times4$ & 24.13 & 1007 & 14.81 & 1242 & 28.15 & 1015 & 23.49 & 1085 & 15.79 & 1057 & 21.27 & 1081 \\
COCOM & $\times4$ & 26.73 & 1007 & 18.83 & 1242 & 23.92 & 1015 & 31.56 & 1085 & 25.60 & 1057 & 25.33 & 1081 \\
\rowcolor[gray]{0.9} 
\multicolumn{14}{c}{\textbf{Plug-in Methods}} \\
xRAG & $\times128$ & 5.93 & 269 & 8.61 & 485 &12.60 & 265 & 17.00 & 324 & 7.77 & 301 & 10.38 & 328 \\
ICAE & $\times128$ & 23.30 & 269 & 18.11 & 485 & 23.57 & 265 & 30.55 & 324 &18.47 & 301 & 22.8 & 328\\
 & $\times16$ & 27.04 & 440 & 19.23 & 667 & 23.97 & 436 & 31.41 & 502 & 23.95 & 476 & 25.12 & 504\\
 & $\times4$ & 27.53 & 1007 & 19.63 & 1242 & 24.51 & 1015 & 31.41 & 1085 & 27.58 & 1057 & 26.13 & 1081\\
ACC-RAG(ours) & adaptive & \underline{41.11} & 630 & \underline{23.33} & 878 & \underline{42.91} & 620 & \underline{33.43} & 689 & \underline{35.37} & 666 & \underline{35.23} & 697 \\
\bottomrule
\end{tabular}
\end{table*}

\subsection{Implementation Details}
\label{sec:imp_detail} We use Mistral-7B-Instruct~\cite{mistral} as the backbone
LLM for all methods, with a decoding temperature of 0 for deterministic
generation. We apply LoRA~\cite{lora} to the model as the compressor, disabling
it during inference as the decoder. All experiments are implemented using PyTorch
and Transformers.
The default retrieval model is ColBERT~\cite{colbert}, using the top-5 ranked documents
for fine-tuning and evaluation. Pretraining is done on single documents while
fine-tuning is performed on five documents. For selector training, we generate
15,000 training samples and 2,000 test samples from the QA train set. The selector
employs a 4-layer/4-head transformer encoder with 256D projection, followed by a
2-layer MLP classifier, incorporating segment embeddings for instruction-context
differentiation and contextual granularity in final features.
More details are provided in Appendix~\ref{sec:append_detail}.
    \section{Experimental Results}
\label{sec:experimental_results}
\subsection{Main Results}
The main results for ACC-RAG are presented in Table~\ref{table:main_results}.
Our ACC-RAG method, trained and inferenced with a granularity sequence of
$[1, 32]$, balances effectiveness and efficiency. Detailed results for other ACC-RAG
configurations are in Section~\ref{subsec:framework}.
The experimental results demonstrate that ACC-RAG significantly (Paired t-test, $p<0.05$) outperforms all compression baselines in M score while maintaining superior efficiency. Notably, under our unified benchmark, most methods (e.g., xRAG, COCOM) exhibit lower effectiveness compared to their original reports, primarily due to their reliance on larger-scale training data and task-specific tuning. In contrast, ACC-RAG achieves a balanced trade-off between effectiveness and efficiency: compared to direct generation, it significantly improves M scores on NQ, TriviaQA, and SQuAD (p<0.05) with minimal drop on WQ and TREC (p>0.05). Meanwhile, it matches or exceeds vanilla RAG’s accuracy on four datasets while reducing FTIT by over $4\times$.

\begin{figure}[t]
    \centering
    \includegraphics[width=\columnwidth]{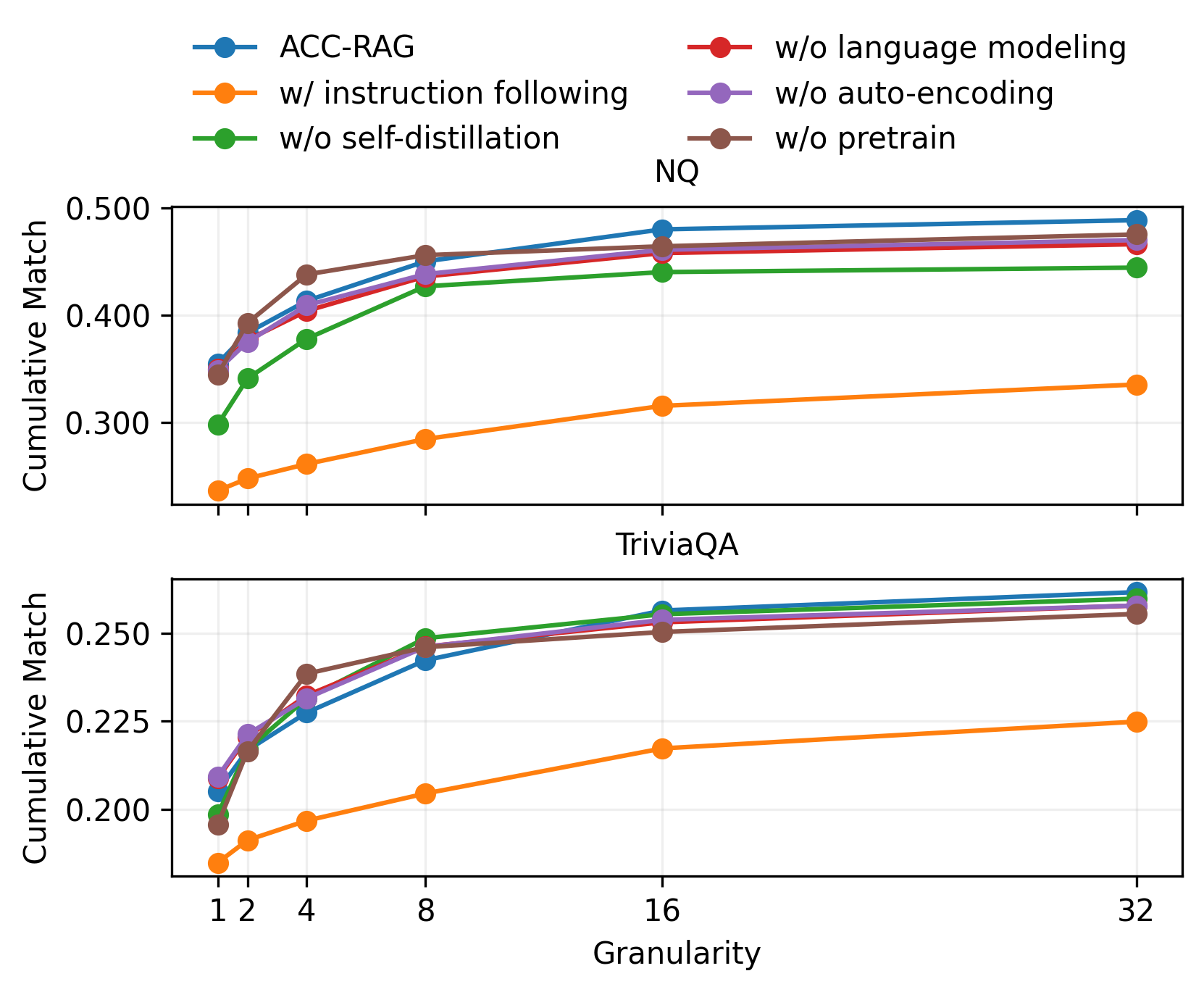}
    \caption{Ablation on different training strategies.}
    \label{figure:compr_loss}
\end{figure}
\subsection{Compressor Analysis}
In this section, we investigate factors influencing compressor performance.
Experiments are conducted under the top-1 retrieval setting for analysis. We
adopt cumulative M scores across different granularities as the evaluation
metric rather than M scores at a single granularity, as the compressor compresses
hierarchical information into position-aware embedding sequences for dynamic aggregation.
Notably, cumulative M scores assume a perfect selector capable of accurately choosing
the optimal granularity for each instruction.
While this does not exist, this section focuses solely on evaluating the compression
capacity of the compressor.

\subsubsection{Compression Rate Analysis}
\label{subsec: compr_rate} The result of different compression rates for compressor
is shown in Appendix~\ref{sec:append_cr}. It can be observed that compressors
with smaller compression rates achieve higher final cumulative M scores, demonstrating
stronger information encoding capabilities. However, compressors with higher compression
rates exhibit accelerated M score accumulation, where the information density
becomes more concentrated at the smaller granularity despite total information
loss.

\subsubsection{Training Strategy Ablation}
\label{subsec: training}
The result of ablation of different training strategies for the compressor is
shown in Figure~\ref{figure:compr_loss}. Our training strategy achieves the highest
cumulative M scores on two datasets, demonstrating its effectiveness. In ablation
experiments, replacing the self-distillation task with instruction following
significantly degraded compression performance. This is likely because instruction
following guides the compressor to encode information into embeddings that do not
align with the decoder’s original generation distribution, impairing the
decoder’s generation ability. This could explain the suboptimal performance of previous
works. Without pretraining, the compressor exhibits faster M-score accumulation at
smaller granularities but slower progression at larger granularities.
Both language modeling and auto-encoding tasks contribute to performance gains, as
their individual removal degrades performance.
The self-distillation proves critical for NQ performance but negligible for
TrivialQA, suggesting its crucial role in certain scenarios.

\begin{figure}[t]
    \centering
    \includegraphics[width=\columnwidth]{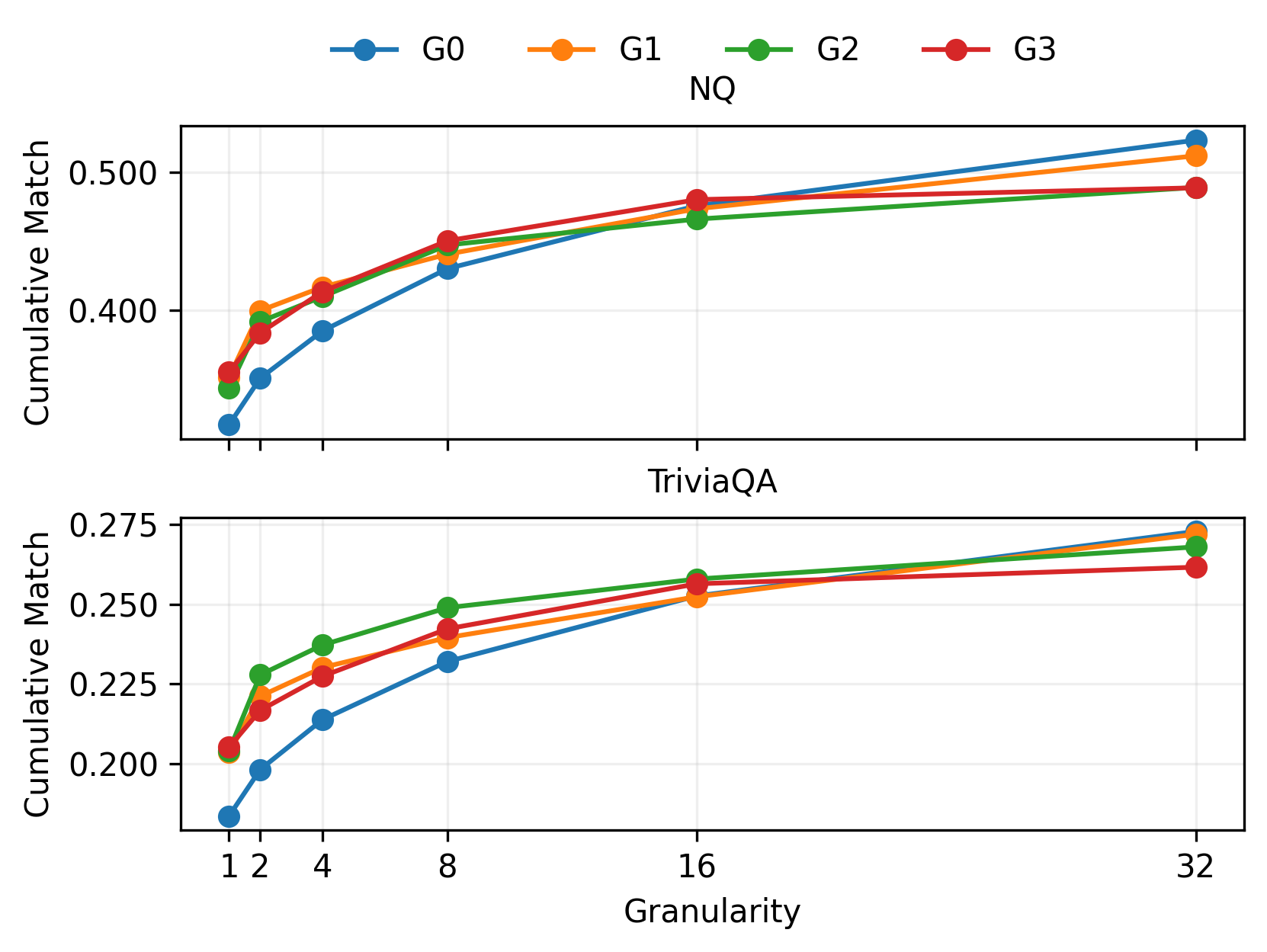}
    \caption{Results of different granularity sequences for compressor. In the legend,
    G0-G3 denote $[32]$, $[1,32]$, $[1,2,8,32]$, $[1,2,4,8,16,32]$ respectively.}
    \label{figure:compr_gran}
\end{figure}


\subsubsection{Granularity Sequence Selection}
\label{subsec: gra_seq}

We investigate the impact of different training granularity sequences on compressor
performance during training. We present results for the inference granularity
sequence $[1,2,4,8,16,32]$, with additional results in Appendix~\ref{sec:append_gran}.
As shown in Figure~\ref{figure:compr_gran}, with the granularity sequence becoming
increasingly dense from G0 to G3, the overall trend indicates a slight decrease
in the final accumulated M scores but accelerated M score accumulation at
smaller granularities.
This suggests that even with a fixed compression rate training approach (i.e.,
the single granularity sequence G0) as in previous work, the compressor retains
hierarchical encoding capabilities. Training with additional smaller granularities
effectively guides the compressor to encode information in the compressed embeddings
at earlier positions. This allows the decoder to access sufficient information
with fewer embeddings, thereby improving inference efficiency. We also observe
that the granularity-based inference method significantly outperforms the single
granularity approach under ideal conditions. The final accumulated M scores on both
datasets notably surpass vanilla RAG, demonstrating the immense potential of our
ACC-RAG framework.

\begin{table}[t]
\centering
\small
\setlength\tabcolsep{3pt}
\caption{
The results of different compressors w/ or w/o a selector during inference.
For each dataset, two columns represent Match (M) and First Token Inference Time (FTIT). G0-G3 denote different granularity sequences.
}
\vspace*{-1.5mm}
\label{table:selector_infer}
\begin{tabular}{c c cc cc}
\toprule
\multirow{2}{*}{\textbf{Compressors}} 
& \multirow{2}{*}{\textbf{Selector}}  
& \multicolumn{4}{c}{\textbf{Datasets}}
\\
\cmidrule(lr){3-6}
&
& \multicolumn{2}{c}{\textbf{NQ}}
& \multicolumn{2}{c}{\textbf{TriviaQA}}
\\ 
\midrule

ACC-G0 & \ding{55} & 40.19 & 1007 & 23.21 & 1242 \\
ACC-G0 & \ding{51} & 41.33 & 786 & 23.38 & 984 \\
\midrule
ACC-G1 & \ding{55} & 40.86 & 1007 & 23.62 & 1242 \\
ACC-G1 & \ding{51} & 41.11 & 630 & 23.33 & 878 \\
\midrule
ACC-G2 & \ding{55} & 39.94 & 1007 & 23.20 & 1242 \\
ACC-G2 & \ding{51} & 39.39 & 622 & 23.07 & 895 \\
\midrule
ACC-G3 & \ding{55} & 41.27 & 1007 & 23.32 & 1242 \\
ACC-G3 & \ding{51} & 40.17 & 631 & 22.58 & 878 \\

\bottomrule
\end{tabular}
\end{table}
\subsection{Selector Analysis}
\subsubsection{Architecture Ablation}
Our ablation study (Appendix~\ref{sec: append_ablation}) reveals that RL serves
as the critical component for sequence-level performance, while supervised learning
exhibits error propagation leading to sequence-level failure due to overemphasis
on individual classification.
In addition, the attention mechanism also plays a key role in enhancing the selector,
while the introduction of segment embeddings and granularity slightly further
improves performance.

\begin{table*}[!t]
\centering
\small
\setlength\tabcolsep{5.5pt}
\caption{
The results of combinations of different compressors and selection strategies. G0-G3 denote different granularity sequences. 
For each dataset, two columns represent Match (M) and Average Context Embedding Counts, with the best metrics bolded and the second-best underlined.
}
\vspace*{-1.5mm}
\label{table:framework}
\begin{tabular}{c c cc cc cc cc cc cc}
\toprule
\multirow{2}{*}{\textbf{Compressors}} 
& \multirow{2}{*}{\textbf{\makecell{Selection\\Strategies}}}  
& \multicolumn{12}{c}{\textbf{Datasets}}
\\
\cmidrule(lr){3-14}
&
& \multicolumn{2}{c}{\textbf{NQ}}
& \multicolumn{2}{c}{\textbf{TriviaQA}}
& \multicolumn{2}{c}{\textbf{WQ}}
& \multicolumn{2}{c}{\textbf{TREC}}
& \multicolumn{2}{c}{\textbf{SQuAD}}
& \multicolumn{2}{c}{\textbf{Average}}
\\ 
\midrule

& G3 & 41.00 & 102 & 23.34 & 105 & 42.32 & 97 & 33.86 & 102 & 35.06 & 109 & 35.12 & 103\\
ACC-G0 & G2 & 41.11 & 104 & \textbf{23.42} & 107 & 42.27 & 99 & \underline{34.15} & 103 & 35.09 & 111 & 35.21 & 105\\
& G1 & \underline{41.33} & 107 & \underline{23.38} & 111 & 42.52 & 104 & 34.01 & 108 
& 35.13 & 114 & \textbf{35.27} & 109\\
\midrule
& G3 & \textbf{41.41} & \underline{56} & 23.33 & \underline{58} & 42.22 & \underline{56} & 33.14 & \underline{57} & 35.16 & \underline{57} & 35.05 & \underline{57}\\
ACC-G1 & G2 & \underline{41.33} & 62 & 23.24 & 63 & \underline{42.67} & 62 & 33.14 & 63 & \underline{35.26} & 62 & 35.13 & 62\\
& G1 & 41.11 & 64 & 23.33 & 64 & \textbf{42.91} & 64 & 33.43 & 64 
& \textbf{35.37} & 64 & \underline{35.23} & 64\\
\midrule
& G3 & 39.42 & \textbf{47} & 22.69 & \textbf{50} & 42.37 & \textbf{48} & 32.71 & \textbf{48} & 35.22 & \textbf{49} & 34.48 & \textbf{48} \\
ACC-G2 & G2 & 39.20 & \underline{56} & 23.11 & 60 & 42.57 & 58 & 33.29 & 58 & 34.97 & 58 & 34.63 & 58\\
& G1 & 39.39 & 63 & 23.07 & 65 & 42.57 & 64 & 33.72 & 64 
& 35.15 & 65 & 34.78 & 64\\
\midrule
& G3 & 40.06 & 63 & 22.67 & 62 & 42.27 & 64 & \underline{34.15} & 64 & 31.51 & 63 & 34.13 & 63\\
ACC-G3 & G2 & 40.17 & 64 & 22.58 & 64 & 42.32 & 64 & \textbf{34.44} & 64 & 31.52 & 64 & 34.21 & 64\\
& G1 & 40.17 & 64 & 22.58 & 64 & 42.32 & 64 & \textbf{34.44} & 64 
& 31.52 & 64 & 34.21 & 64\\

\bottomrule
\end{tabular}
\end{table*}
\subsubsection{Inference Verification}
To validate the effectiveness of the selector during actual inference, we compare
different compressors with and without the selector using the inference granularity
sequence $[1, 32]$. As shown in Table~\ref{table:selector_infer}, ACC-G0 and ACC-G1
with the selector outperformed the versions without the selector on nearly all datasets
with superior efficiency (with FTIT reduced by over 20\%). ACC-G2 and ACC-G3 with
the selector also maintain comparable performance to their counterparts without the
selector, while achieving over 30\% FTIT reduction.

\subsection{Framework Evaluation}
\label{subsec:framework}

The results of combinations of compressors trained with different granularity
sequences and inference selection strategies are shown in Table~\ref{table:framework}.
We observe that, for the same compressor, sparser inference granularity sequences
yield higher M scores but require more context embeddings. This is due to
sparser sequences reducing classification errors by the selector, while potentially
missing more precise termination points. Among all compressors, ACC-G0+G1 achieves
the highest average M score, while ACC-G1+G1 reaches a similar M score with less
than 60\% of the context embeddings, demonstrating the effectiveness of the
multi-granularity training strategy in hierarchical information encoding. Additionally,
training the compressor and selection strategy with denser granularity sequences
further enhances efficiency, as exemplified by ACC-G2+G3, which optimized
inference efficiency at the expense of a slight M score decrease.

\subsection{Case Study}
We present a representative case in Appendix~\ref{sec:append_case}, which demonstrates
the outstanding performance of ACC-RAG in both efficiency and effectiveness.
\begin{table}[t]
\centering
\small
\setlength\tabcolsep{5.5pt}
\caption{
The results of RAG performance between vanilla RAG and ACC-RAG. 
For each dataset, two columns represent resilience and boost rate, with the highest score bolded and the second-highest underlined.
}
\vspace*{-1.5mm}
\label{table:rag}
\begin{tabular}{c cc cc}
\toprule
\multirow{2}{*}{\textbf{Method}} 
& \multicolumn{4}{c}{\textbf{Datasets}}
\\
\cmidrule(lr){2-5}
& \multicolumn{2}{c}{\textbf{NQ}}
& \multicolumn{2}{c}{\textbf{TriviaQA}}
\\ 
\midrule

RAG & 72.98 & \textbf{28.78} & 82.06 & \textbf{8.97} \\
ACC-RAG-G0 & 76.18 & \underline{22.94} & 83.22 & \underline{7.91} \\
ACC-RAG-G1 & \textbf{77.95} & 21.67 & \underline{83.56} & 7.75 \\
ACC-RAG-G2 & 74.82 & 20.69 & 83.22 & 7.52 \\
ACC-RAG-G3 & \underline{77.79} & 20.31 & \textbf{83.95} & 6.72 \\

\bottomrule
\end{tabular}
\end{table}
\subsection{Scalability Evaluation}
We observe an excellent trade-off between effectiveness and efficiency with
Llama3-3B-Instruct and Llama3-8B-Instruct (results in Appendix~\ref{sec:append_scal}).
ACC-RAG achieves results comparable to RAG on both models and datasets (with <10\%
M score difference), while improving speed by 4 to 5 times (with FTIT reduced by
76\% to 83\%), fully showcasing the scalability of the method.

\subsection{RAG Integration Analysis}
We evaluate the RAG performance using the resilience rate and the boost rate proposed
in \cite{xrag}. The resilience rate quantifies the proportion of correct decoder
responses retained post-retrieval, while the boost rate measures the percentage of
initially incorrect responses corrected through retrieval augmentation.
As shown in Table~\ref{table:rag}, RAG exhibits a higher boost rate, while ACC-RAG
demonstrates a higher resilience rate. As the training granularity sequence for compressor
becomes denser, the overall trend is an increase in the boost rate and a
decrease in the resilience rate.

\subsection{OOD Analysis}
We evaluate the Out of Distribution(OOD) performance of our method(results in Appendix~\ref{sec:append_ood}), demonstrating its exceptional generalization ability to unseen supporting documents and unseen queries from different domains.

\subsection{Additional Computation Analysis}
Our approach incurs additional computational overhead during inference due to the introduced selector. Analysis in Appendix~\ref{sec:append_ac} demonstrates the favorable trade-off between controlled computational cost and substantial performance gains.
    \section{Conclusion}
\label{sec:conclusion}
In this paper, we propose Adaptive Context Compression for RAG, a framework that dynamically adjusts compressed embeddings required during inference.
The framework effectively improves inference efficiency while maintaining the quality benefits of retrieval.
Experimental results under standardized benchmark show that our method achieves the best results under context compression and comparable or superior accuracy with over 4× inference efficiency compared to RAG.
    \section*{Limitations}
\label{sec:limitations}
Despite the promising results demonstrated in this paper, there are several limitations to our framework.
\begin{itemize}
    \item We have only conducted a preliminary exploration of the selector and its performance is currently the largest bottleneck in the entire framework. 
    Improving the prediction accuracy of the selector is key to further unlocking the potential of adaptive text compression.
    \item Our work focuses on adaptive context selection, lacking experimental analysis under controlled conditions, which remains a promising direction for future research.
    \item We have trained the compressor and selector separately, without further exploring the potential of joint training. This may hinder the alignment between the two modules, potentially reducing the overall performance of the framework.
    \item Due to computational constraints, we have not explored the performance of ACC-RAG on larger models and longer texts, which may limit its applicability in certain scenarios.
\end{itemize}
    \section*{Ethics Statement}
\label{sec:ethics}
The research conducted in this paper centers around the improvement of efficiency in retrieval augmented generation. Our framework accelerates the inference of retrieval augmented models by compressing long context into compact embeddings and dynamically selecting them. 

We acknowledge the importance of the ACM Code of Ethics and affirm our adherence to it. We ensure that this work is compatible with the provided code. All data used in this study is obtained from existing benchmarks, while all models used are publicly available, thus ensuring a high level of transparency and reproducibility in our experimental procedure.

We have made every effort to ensure that our research does not harm individuals or groups, nor does it involve any form of deception or potential misuse of information.

    \bibliography{references}
    \clearpage
\appendix

\section{Datasets}
\label{sec:append_dataset}
We use the dataset version processed by DPR\footnote{\url{https://github.com/facebookresearch/DPR}} including retrieval corpus and five QA datasets, which is licensed by \textit{CC-BY-NC 4.0}.
All datasets are English-language resources within the question answering (QA) domain.
The datasets we used inherit from them while maintaining consistency in QA intent.
We ensure that the dataset is free from harmful content or information leakage through a sampling approach.
The detailed statistics of the dataset are presented in Table~\ref{tab:dataset_statistics}. Notably, based on whether the supporting documents are present in the pretraining dataset, the development set of the fine-tuning data is further divided into seen and unseen subsets. 
\begin{table}[H]
\centering
\setlength\tabcolsep{4pt}
\begin{tabular}{l c cc c}
\toprule
\textbf{Datasets}
& \textbf{Train}
& \multicolumn{2}{c}{\textbf{Dev}}
& \textbf{Test}
\\
\midrule
Natural Questions & 58880 & 3195 & 3320 & 3610 \\
TriviaQA & 60413 & 2102 & 4658 & 11313 \\
WebQuestions & 2474 & 77 & 201 & 2032 \\
CuratedTREC & 1125 & 32 & 84 & 694 \\
SQuAD v1.1 & 70096 & 6985 & 936 & 10570 \\
\bottomrule
\end{tabular}
\caption{Number of QA pairs in each dataset  across Train, Dev, and Test splits. The two columns of Dev denote the seen and the unseen subsets respectively.}
\label{tab:dataset_statistics}
\end{table}

\section{Models}
We use Mistral-7B-Instruct-v0.2\footnote{\url{https://huggingface.co/mistralai/Mistral-7B-Instruct-v0.2}}, Llama3.2-3B-Instruct\footnote{\url{https://huggingface.co/meta-llama/Llama-3.2-3B-Instruct}} and Llama3.1-8B-Instruct\footnote{\url{https://huggingface.co/meta-llama/Llama-3.1-8B-Instruct}} as our backbone LLM.
The Mistral model is licensed by \textit{Apache-2.0} and two Llama models are licensed by \textit{Meta Llama3 Community License}.
The retriever we used is ColBERT-v2\footnote{\url{https://github.com/stanford-futuredata/ColBERT}} with \textit{MIT License}.

\section{Implementation Details}
\label{sec:append_detail}

In Table~\ref{tab:pretraining} and Table~\ref{tab:finetune}, we list the hyperparameters for compressor pretraining and fine-tuning.
These parameters are primarily inherited from prior work (e.g., learning rate, LoRA configuration), as the same backbone LLM and similar training tasks are used. Additionally, all methods are trained under the same configuration for a fair comparison, ensuring no intentional optimization bias towards our approach.
For efficient training, at each training step, we randomly sample only one granularity from granularity sequence and one task to optimize during pretraining.
In our configuration, we pretrain the compressor with roughly 20 hours and finetune with roughly 50 hours.
For selector training, we list the hyperparameters in Table~\ref{tab:selector_train}.
We primarily search the hyperparameters of learning rates and find that a range between $2e^{-4}$ and $1e^{-4}$ yields the best results. Additionally, we adjusted the number of attention heads and layers in the transformer encoder. We observed that increasing both parameters beyond 4 provided no improvement, while reducing them below 4 resulted in a slight decline in performance.
The entire training process took approximately 1 hour, and we ultimately selected the model with the highest sequence accuracy on the test set.
All experiments are implemented using PyTorch\footnote{\url{https://pytorch.org/}} and Hugging Face Transformers\footnote{\url{https://github.com/huggingface/transformers}}.

\begin{table}[t]
\centering
\setlength\tabcolsep{6.5pt}
\begin{tabular}{c c}
\toprule
\textbf{Hyperparameter}
& \textbf{Assignment}
\\
\midrule
optimizer & AdamW \\
learning rate & $1e^{-4}$ \\
lr scheduler type & linear \\
warmup ratio & 0.03 \\
weight decay & 0.0 \\
epochs & 1 \\
flash attention & True \\
batch size & 8 \\
gradient accumulation steps & 4 \\
LoRa $r$ & 128 \\
LoRa alpha & 32 \\
LoRa dropout & 0.05 \\
LoRa bias & None \\
gpu & $1\times$A100 80G \\
max sequence length & 320 \\
\bottomrule
\end{tabular}
\caption{Hyperparameters for compressor pretraining.}
\label{tab:pretraining}
\end{table}
\begin{table}[t]
\centering
\setlength\tabcolsep{6.5pt}
\begin{tabular}{c c}
\toprule
\textbf{Hyperparameter}
& \textbf{Assignment}
\\
\midrule
optimizer & AdamW \\
learning rate & $1e^{-5}$ \\
lr scheduler type & linear \\
warmup ratio & 0.03 \\
weight decay & 0.0 \\
epochs & 5 \\
flash attention & True \\
batch size & 2 \\
gradient accumulation steps & 4 \\
LoRa $r$ & 128 \\
LoRa alpha & 32 \\
LoRa dropout & 0.05 \\
LoRa bias & None \\
gpu & $1\times$A6000 48G \\
max sequence length & 1024 \\
\bottomrule
\end{tabular}
\caption{Hyperparameters for compressor fine-tuning.}
\label{tab:finetune}
\end{table}
\begin{table}[t]
\centering
\setlength\tabcolsep{6.5pt}
\begin{tabular}{c c}
\toprule
\textbf{Hyperparameter}
& \textbf{Assignment}
\\
\midrule
optimizer & AdamW \\
learning rate & $1e^{-4}$ \\
lr scheduler type & linear \\
warmup ratio & 0.0 \\
weight decay & 0.0 \\
epochs & 50 \\
batch size & 128 \\
gradient accumulation steps & 1 \\
gpu & $1\times$A6000 48G \\
\bottomrule
\end{tabular}
\caption{Hyperparameters for selector training.}
\label{tab:selector_train}
\end{table}

\begin{figure}[htbp]
 \centering
 \includegraphics[width=\columnwidth]{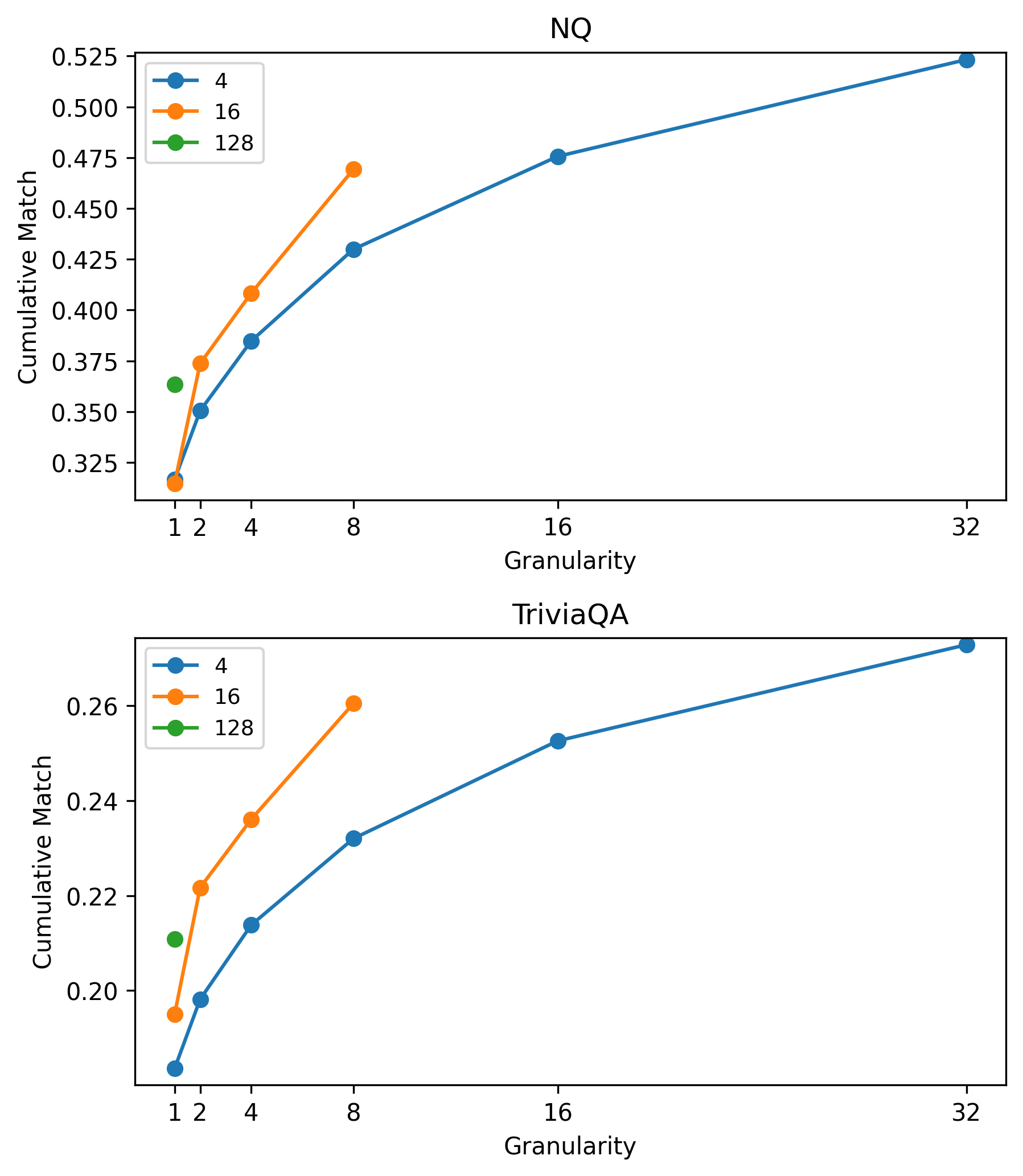}
 \caption{Ablation on different compression rates for compressor.}
 \label{figure:compr_cr}
\end{figure}

\begin{figure}[t]
 \centering
 \includegraphics[width=\columnwidth]{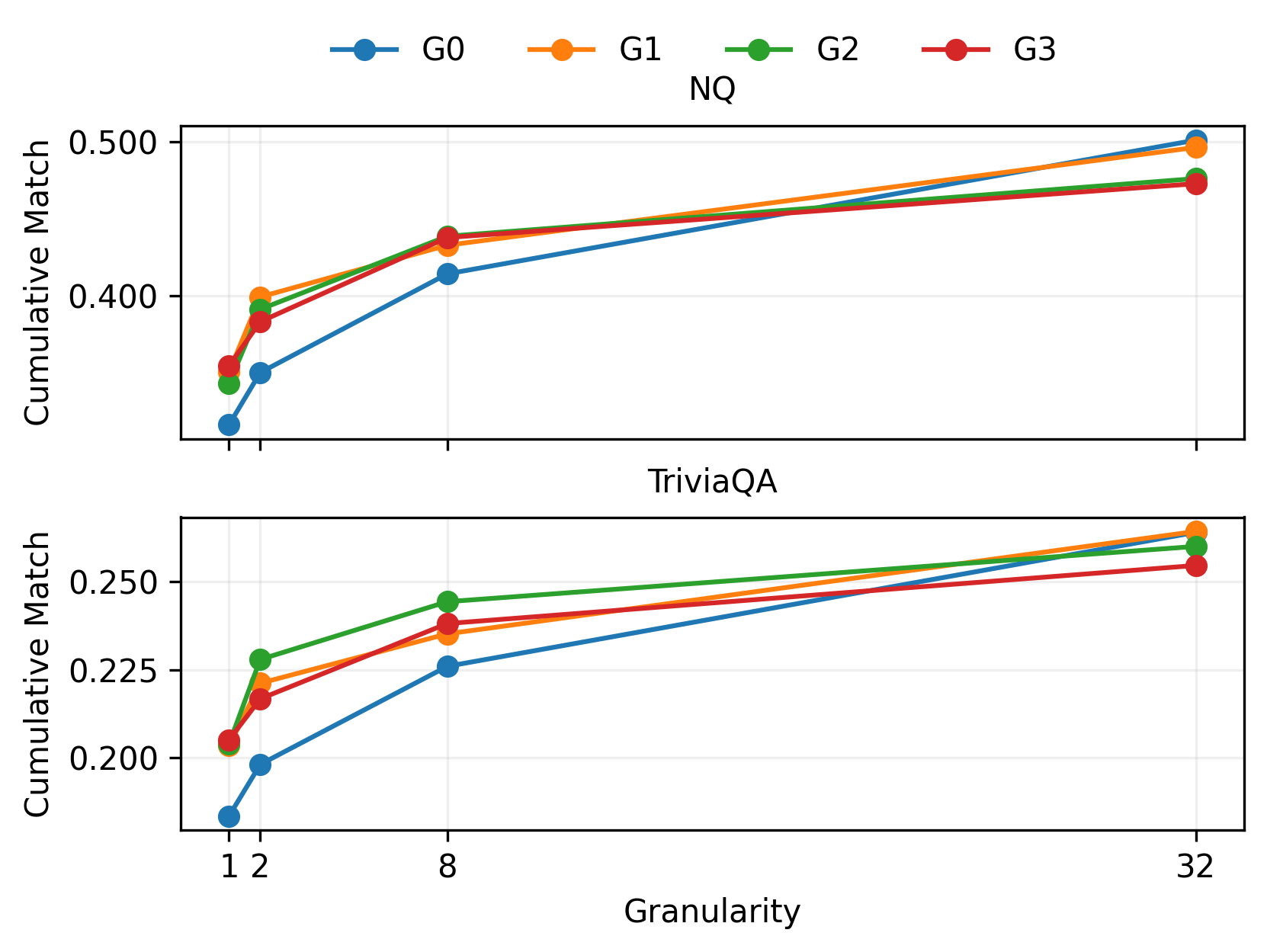}
 \caption{Results of different granularity sequences for compressor under inference granularity sequence $[1,2,8,32]$. In the legend, G0-G3 denote $[32]$, $[1,32]$, $[1,2,8,32]$, $[1,2,4,8,16,32]$ respectively.}
 \label{figure:compr_gran_4}
\end{figure}

\begin{figure}[t]
 \centering
 \includegraphics[width=\columnwidth]{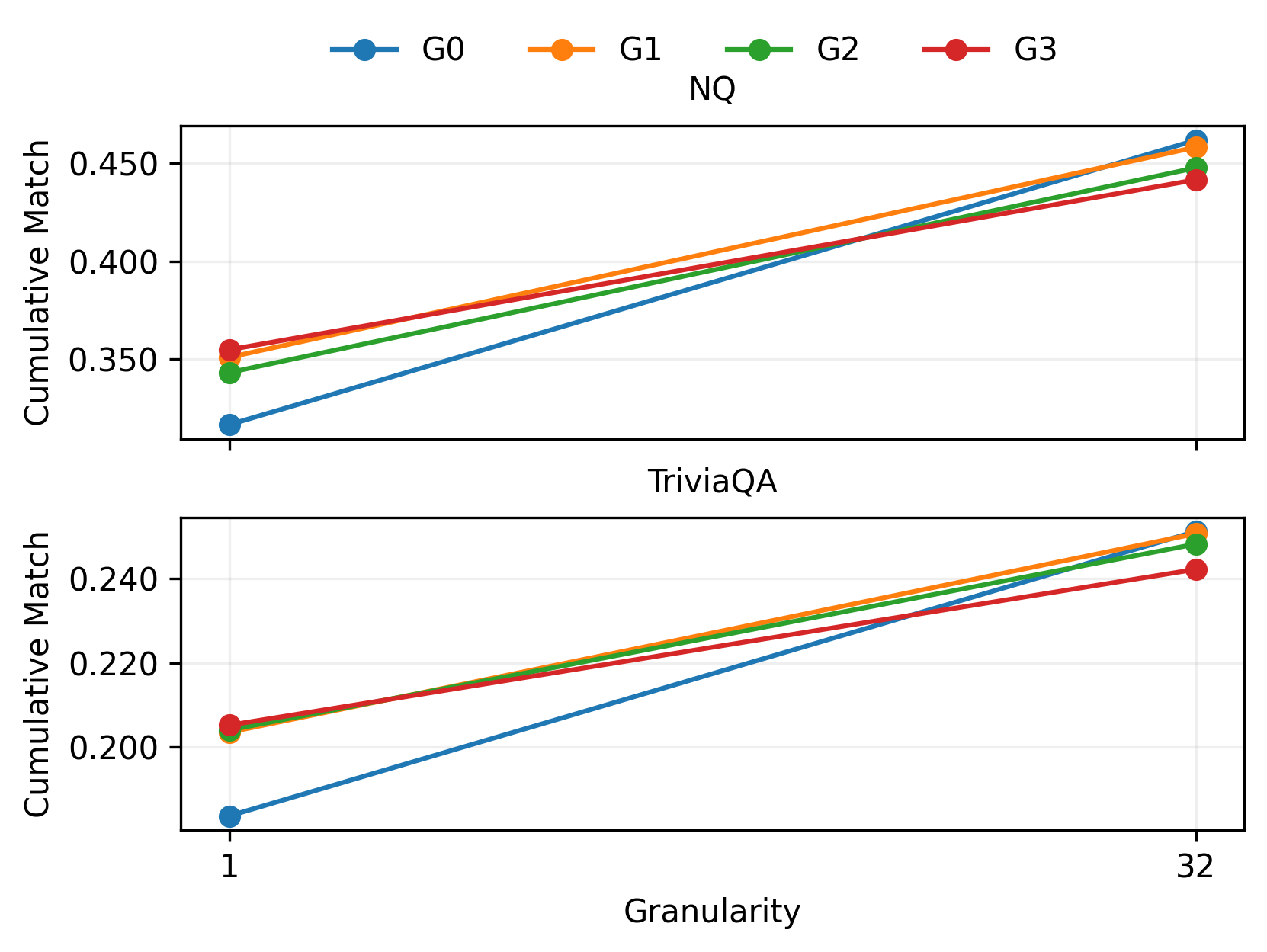}
 \caption{Results of different granularity sequences for compressor under inference granularity sequence $[1,32]$. In the legend, G0-G3 denote $[32]$, $[1,32]$, $[1,2,8,32]$, $[1,2,4,8,16,32]$ respectively.}
 \label{figure:compr_gran_2}
\end{figure}

\section{Analysis on Compression Rate of Compressor}
\label{sec:append_cr}
We investigate the impact of different compression rates on compressor performance during training.
The results are shown in Figure~\ref{figure:compr_cr}.
We train the model with a single granularity to facilitate analysis, where the granularity is the count of compressed embeddings obtained by compression at a specific compression rate.
\begin{table*}[!t]
\centering
\small
\setlength\tabcolsep{4.5pt}
\caption{
The OOD performance of unseen supporting documents.
The highest Match(M) and lowest First Token Inference Time(FTIT) is bolded.
}
\vspace*{-1.5mm}
\label{table:ood-doc}
\begin{tabular}{c c c c c c c c}
\toprule
\multirow{2}{*}{\textbf{Method}} 
& \multicolumn{7}{c}{\textbf{Datasets}}
\\
\cmidrule(lr){2-8}
& \textbf{NQ}
& \textbf{TriviaQA}
& \textbf{WQ}
& \textbf{TREC}
& \textbf{SQuAD}
& \textbf{Average}
& \textbf{Average FTIT}
\\ 
\midrule

RAG & 69.97 & \textbf{27.44} & 49.75 & \textbf{39.29} & \textbf{59.62} & \textbf{49.21} & 3268 \\
ICAE($\times4$) & 61.66 & 26.26 & 50.25 & 35.71 & 47.97 & 44.37 & 1052 \\
ACC-RAG(ours) & 63.34 & 26.56 & \textbf{50.75} & \textbf{39.29} & 51.60 & 46.31 & \textbf{673} \\

\bottomrule
\end{tabular}
\end{table*}

\section{Analysis on Granularity Sequences of Compressor}
\label{sec:append_gran}
The results of different granularity sequences for compressor under inference granularity sequence $[1,2,8,32]$ and $[1,32]$ are shown in Figure~\ref{figure:compr_gran_4} and Figure~\ref{figure:compr_gran_2}. 
The conclusions are consistent with Section~\ref{subsec: gra_seq}.
We observe that the final accumulated M scores exceeds the vanilla RAG even with only two inference granularities.

\section{Ablation Study of Selector}
\label{sec: append_ablation}
We investigate the impact of different components of selector for classification performance.
The default setting of ACC-Selector is described in Section~\ref{sec:imp_detail}.
We adopt the sequence classification accuracy as evaluation metric, verifying whether the selector’s first prediction of "sufficient" is correct.
We report the results in test set in Table~\ref{tab:selector}.
\begin{table}[H]
\centering
\setlength\tabcolsep{4pt}
\begin{tabular}{c c}
\toprule
\textbf{Methods}
& \textbf{Sequence Accuracy}
\\
\midrule
ACC-Selector &  70.75 \\
w/o RL & 64.25 \\
w/o attention & 66.30 \\
w/o segment embeddings & 69.35 \\
w/o granularity & 69.50\\
\bottomrule
\end{tabular}
\caption{Ablation on selector.}
\label{tab:selector}
\end{table}

\section{Representative Case}
\label{sec:append_case}
To provide a more intuitive understanding of the effectiveness of ACC-RAG, we present a representative case in Table~\ref{tab:case}. 
In this case, direct generation and ACC-RAG-G2 with a fixed compression rate of 16 fail to answer the question. 
However, our method ACC-RAG-G2 with inference granularity sequence $[1,32]$ not only provides the correct answer but also uses the fewest context tokens compared to ACC-RAG-G2 with a fixed compression rate of 4 and RAG.

\section{Scalability Analysis}
\label{sec:append_scal}
To investigate whether ACC-RAG is applicable to other models, we apply our approach to both Llama3-3B-Instruct and Llama3-8B-Instruct models and report the results shown in Table~\ref{table:scalability}.
\begin{table}[H]
\centering
\small
\setlength\tabcolsep{2.5pt}
\caption{
The results of ACC-RAG in Llama3 for scalability study.
For each dataset, two columns represent Match (M) and First Token Inference Time (FTIT).
}
\vspace*{-1.5mm}
\label{table:scalability}
\begin{tabular}{c c cc cc}
\toprule
\multirow{2}{*}{\textbf{Model}} 
& \multirow{2}{*}{\textbf{Method}}  
& \multicolumn{4}{c}{\textbf{Datasets}}
\\
\cmidrule(lr){3-6}
&
& \multicolumn{2}{c}{\textbf{NQ}}
& \multicolumn{2}{c}{\textbf{TriviaQA}}
\\ 
\midrule

\multirow{3}{*}{Llama3-3B} 
& - & 36.37 & 65 & 15.33 & 101 \\
& RAG & 40.72 & 1037 & 22.26 & 966 \\
& ACC-RAG & 38.76 & 180 & 18.62 & 225 \\
\midrule

\multirow{3}{*}{Llama3-8B} 
& - & 39.72 & 139 & 20.56 & 235 \\
& RAG & 44.43 & 2169 & 23.64 & 2089 \\
& ACC-RAG & 41.32 & 380 & 21.19 & 489 \\

\bottomrule
\end{tabular}
\end{table}

\section{OOD Analysis}
\label{sec:append_ood}
To validate the OOD capability of our method, we conduct evaluations from two perspectives: (1) unseen supporting documents and (2) unseen queries from different domains. The former is evaluated on the unseen Devset where each case utilizes a ground-truth supporting document absent from the pretraining dataset. Results in Table~\ref{table:ood-doc} demonstrate that our method outperforms the strongest baseline in both effectiveness and efficiency. Compared to Vanilla RAG, our approach achieves over 4× faster inference speed while maintaining comparable accuracy across three datasets, with superior performance observed on two datasets. The latter evaluation employs additional dataset from two distinct tasks (HotpotQA~\cite{HotpotQA} and FEVER~\cite{FEVER}). As shown in Table~\ref{table:ood-domain}, our method again demonstrates superior performance compared to strongest baseline while maintaining the 4× speed improvement over Vanilla RAG without compromising accuracy on both datasets.
\begin{table}[t]
\centering
\small
\setlength\tabcolsep{5.5pt}
\caption{
The OOD performance on unseen queries from different domains. 
For each dataset, two columns represent Match (M) and First Token Inference Time (FTIT).
}
\vspace*{-1.5mm}
\label{table:ood-domain}
\begin{tabular}{c cc cc}
\toprule
\multirow{2}{*}{\textbf{Method}} 
& \multicolumn{4}{c}{\textbf{Datasets}}
\\
\cmidrule(lr){2-5}
& \multicolumn{2}{c}{\textbf{HotpotQA}}
& \multicolumn{2}{c}{\textbf{FEVER}}
\\ 
\midrule

LLM & 33.70 & 290 & 74.44 & 293 \\
RAG & 44.96 & 3313 & 83.60 & 3325 \\
ICAE($\times4$) & 40.96 & 1114 & 79.62 & 1120 \\
ACC-RAG(ours) & 41.98 & 689 & 81.55 & 693\\

\bottomrule
\end{tabular}
\end{table}

\section{Additional Computation Analysis}
\label{sec:append_ac}
On the NQ dataset, our method achieves an average FTIT of 630, with the selector accounting for 47 (\textasciitilde7\% of total time). This selective mechanism enables our method to outperform RAG (3264 FTIT) and fixed-ratio compression approaches with same compression rate (1007 FTIT).

\section{Potential Risks}
Our method reduces RAG input length while preserving contextual integrity, yet inherits inherent potential risks of RAG such as retrieval-related risks (data bias/information leakage), misleading outputs from outdated/incorrect knowledge, adversarial database manipulation leading to harmful responses and so on.
In addition, although not observed at present, compressing text into embeddings may potentially distort the original information, leading to harmful outputs generated by the model.

\begin{table}[!]
\centering
\setlength\tabcolsep{4pt}
\small
\begin{tabular}{p{1\columnwidth}}
\toprule
\multicolumn{1}{c}{\textbf{Model Input}} \\
\midrule
\textbf{Question:} what's the official symbol of the carnival of quebec? \\
\textbf{Context:}\\
1. It is adorned by an arrowed pattern and was worn around the winter coats of the time. It is also a symbol of the Lower Canada Rebellion and the Quebec Winter Carnival, as it is worn by the festival mascot, \textcolor{blue}{\textbf{Bonhomme Carnaval}}. Imitations are sold and seen throughout the carnival. The belt is represented in a number of artistic creations, such as the illustration ``Le Vieux de '37'' by Henri Julien, the painting ``L'Assemblée des six-comtés'' by Charles Alexander Smith and the song ``Mon Pays, suivi du Reel des Aristocrates'' from néo-trad musical band Les Cowboys Fringants. \\
2. Carifiesta Carifiesta () is an annual Caribbean Carnival held in Montreal, Quebec, Canada. It was established in 1974, and is held in July. The event is coordinated by the ``Caribbean Cultural Festivities Association'', a nonprofit organization. Carifiesta was established prior to some Carnivals that take place in the Caribbean, Cayman Carnival Batabano for example.\\
3. Quebec Winter Carnival The Quebec Winter Carnival (), commonly known in both English and French as Carnaval, is a pre-Lenten festival held in Quebec City. After being held intermittently since 1894, the ``Carnaval de Québec'' has been celebrated annually since 1955. That year ``\textcolor{blue}{\textbf{Bonhomme Carnaval}}'', the mascot of the festival, made his first appearance. \\
4. Canadian rule (maple leaves) and with Quebec's motto below ``Je me souviens''. Je me souviens (``I remember'') was first carved under the coat of arms of Quebec's Parliament Building façade in 1883. It is an official part of the coat of arms and has been the official licence plate motto since 1978, replacing ``La belle province'' (the beautiful province). The expression ``La belle province'' is still used mostly in tourism as a nickname for the province. \\
5. Montreal Carnivale. The carnivale is named after its mascot Bon Soo, created by Ken MacDougall. Bon Soo Winter Carnival The Bon Soo Winter Carnival is an annual winter carnival in Sault Ste. Marie, Ontario, Canada. The carnival, held every February, began in , and has grown to become one of Ontario's top 50 festivals as well as Canada's second largest winter carnival next to the Montreal Carnivale in Quebec.\\
\midrule
\multicolumn{1}{c}{\textbf{Model Outputs}} \\
\midrule
\textbf{Direct Generation:}
The official symbol of the Carnival of Quebec, also known as Carnaval de Québec, is a snowgoose. \\
\textcolor{red}{\textbf{Generated with 0 context tokens (incorrect answer).}}\\
\textbf{ACC-RAG-G2 with fixed compression rate 16:}
The official symbol of the Carnival of Quebec is the ``Fleur-de-lis'' or ``leur-de-lis carnival'' as it is commonly known.  \\
\textcolor{red}{\textbf{Generated with 40 context tokens, wrong answer.}}\\
\textbf{ACC-RAG-G2(ours):}
The official symbol of the Carnival of Quebec is ``Bonhomme Carnaval,'' a jovial snowman-like figure wearing a red hat and scarf, who is the mascot of the Quebec Winter Carnival.  \\
\textcolor[HTML]{006400}{\textbf{Generated with 64 context tokens, right answer.}}\\
\textbf{ACC-RAG-G2 with fixed compression rate 4:}
The official symbol of the Carnival of Quebec is ``Bonhomme Carnaval,'' a mascot character with a red nose, a red hat, and a red scarf, who is depicted as a snowman.  \\
\textcolor[HTML]{006400}{\textbf{Generated with 160 context tokens, right answer.}}\\
\textbf{Vanilla RAG:}
The official symbol of the Quebec Winter Carnival is ``Bonhomme Carnaval'' and the maple leaf with the motto ``Je me souviens'' (I remember).  \\
\textcolor[HTML]{006400}{\textbf{Generated with 640 context tokens, right answer.}}\\
\bottomrule
\end{tabular}
\caption{A Case of generated responses using different methods.}
\label{tab:case}
\end{table}
\end{document}